\documentclass[runningheads]{llncs}
\usepackage{graphicx}
\usepackage{dsfont}
\usepackage{tikz}
\usepackage{comment}
\usepackage{amsmath,amssymb} % define this before the line numbering.
\usepackage{color}
\usepackage{svg}
\usepackage{authblk}
\usepackage{enumitem}
\usepackage{multirow}
\usepackage{breakcites}

\begin{document}
% \linenumbers
\pagestyle{headings}
\mainmatter
\def\ECCVSubNumber{34}  % Insert your submission number here

\newcommand\boldred[1]{\textcolor{red}{\textbf{#1}}}
\makeatletter
\newcommand{\printfnsymbol}[1]{%
  \textsuperscript{\@fnsymbol{#1}}%
}
\title{Adaptive Mask-based Pyramid Network for Realistic Bokeh Rendering}
% Replace with your title

\begin{comment}
\titlerunning{ECCV-22 submission ID \ECCVSubNumber} 
\authorrunning{ECCV-22 submission ID \ECCVSubNumber} 
\author{Anonymous ECCV submission}
\institute{Paper ID \ECCVSubNumber}
\end{comment}
%******************

% \begin{comment}
\titlerunning{Adaptive Bokeh Rendering}

\author{Konstantinos Georgiadis\thanks{The authors have contributed equally.}\inst{2}\and
Albert Sa\`a-Garriga\printfnsymbol{1} \inst{1}\and
Mehmet Kerim Yucel\printfnsymbol{1} \inst{1}\and
Anastasios Drosou \inst{2}\and
Bruno Manganelli \inst{1}}
\authorrunning{Georgiadis et al.}
%03/05 – We heard reports of a citation block overflow issue with the ECCV submission template: Citation blocks in square brackets sometimes overflow onto the margins instead of wrapping to the next line. This seems to be an old issue with the template and there may be different ways of solving it. One simple fix is adding the instruction “\sloppy” to the LaTeX preamble. Note that this solution is offered for the review version; camera ready may require removal of this instruction.

\institute{Samsung Research UK \and
Centre for Research and Technology Hellas (CERTH), Information Technologies Institute, Thessaloniki, Greece
}
% \end{comment}
%******************
\maketitle

\begin{abstract}
Bokeh effect highlights an object (or any part of the image) while blurring the rest of the image, and creates a visually pleasant artistic effect. Due to the sensor-based limitations on mobile devices, machine learning (ML) based bokeh rendering has gained attention as a reliable alternative. In this paper, we focus on several improvements in ML-based bokeh rendering; i) on-device performance with high-resolution images, ii) ability to guide bokeh generation with user-editable masks and iii) ability to produce varying blur strength. To this end, we propose Adaptive Mask-based Pyramid Network (AMPN), which is formed of a Mask-Guided Bokeh Generator (MGBG) block and a Laplacian Pyramid Refinement (LPR) block. MGBG consists of two lightweight networks stacked to each other to generate the bokeh effect, and LPR refines and upsamples the output of MGBG to produce the high-resolution bokeh image. We achieve i) via our lightweight, mobile-friendly design choices, ii) via the stacked-network design of MGBG and the weakly-supervised mask prediction scheme and iii) via manually or automatically editing the intensity values of the mask that guide the bokeh generation. In addition to these features, our results show that AMPN produces competitive or better results compared to existing methods on the EBB! dataset, while being faster and smaller than the alternatives.

\end{abstract}

%%%%%%%%% BODY TEXT

\section{Introduction}

Bokeh effect is one of the  fundamental photography techniques, where an object (or a region) in the image is effectively highlighted by blurring out the rest of the image. Traditionally, such an effect is achieved via focusing the camera on an area and taking the photo using a wide aperture lens. Although it is achievable with appropriate cameras with fast lenses with large apertures, or even with mobile devices with stereo setups, bokeh rendering may not be feasible for all mobile devices due to sensory/hardware constraints. Synthetically generating the bokeh effect through machine-learning (ML) based methods are therefore viable alternatives, especially for setups without adequate hardware.

There has been a rising interest in synthetic bokeh rendering, especially following the release of EBB! dataset \cite{ignatov2020rendering} and bokeh rendering challenges \cite{ignatov2019aim,ignatov2020aim}. Starting with earlier methods focusing on portrait images \cite{shen2016automatic,shen2016deep}, later methods leveraged advances in image-to-image methods and generative models \cite{ignatov2020rendering,dutta2021stacked,choi2020efficient,purohit2019depth,qian2020bggan,peng2022bokehme,zheng2022constrained,luo2020bokeh,nagasubramaniam2022bokeh,wang2018deeplens,dutta2021depth}. Depth/disparity \cite{wang2018deeplens,purohit2019depth,ignatov2020rendering,dutta2021depth} and saliency maps \cite{purohit2019depth,ignatov2019aim}, as well as segmentation maps \cite{zhu2017fast} have found use as the principal guiding component for bokeh rendering, however, such methods require additional ground-truth information during training and require additional components in inference time. Conversely, methods not using any guidance at all \cite{dutta2021stacked,nagasubramaniam2022bokeh,qian2020bggan,choi2020efficient} are essentially learning an inflexible mapping, where in-focus areas will be implicitly learned by the network and can not be changed by the user. Furthermore, ML-based methods essentially fit to the blur strength of the training data and can generate only fixed bokeh styles \cite{peng2022bokehme}, leading to limited user interaction.

Thinking from a mobile use case standpoint, we focus on three areas of improvement; i) fast on-device runtimes with high-resolution images, ii) ability to guide bokeh generation with user-editable masks without additional compute in inference time and iii) ability to render bokeh effect with varying blur strengths. There are fast, on-device methods for bokeh rendering \cite{choi2020efficient,dutta2021depth,dutta2021stacked,ignatov2020rendering}, but none of them meets the criteria ii) and iii). There are methods leveraging external guidance for bokeh rendering (i.e. criteria ii)) \cite{wang2018deeplens,purohit2019depth,ignatov2020rendering,dutta2021depth,ignatov2019aim,zhu2017fast}, but they require additional processing to generate such guidance (i.e. depth models, saliency models, etc). A recent work \cite{peng2022bokehme} focuses on the ability to generate varying blur strengths (criteria iii)), but their solution is not tailored for on-device performance and therefore not suitable for mobile use cases.

In light of these above mentioned criteria, we propose Adaptive Mask-based Pyramid Network (AMPN) for mobile-friendly, realistic bokeh rendering. AMPN is formed of two main building blocks; i) Mask-Guided Bokeh Generator (MGBG) and ii) Laplacian Pyramid Refinement (LPR). MGBG consists of two stacked networks, where the first one is responsible for mask prediction and the other one is for bokeh generation, guided by the predicted mask. LPR essentially refines the low-resolution bokeh output of MGBG and upsamples it to the input resolution. Using lightweight designs in both blocks lets us operate efficiently, whereas LPR lets us produce high fidelity, high-resolution outputs without the need of third party solutions for upsampling/super-resolution (meeting criteria i). Furthermore, if a user is providing its mask, we can completely remove the mask prediction network during inference, making our pipeline even more compact.

Having strong user-guidance often requires strong supervision (i.e. ground-truths) in (guidance) mask prediction, which may not be available in every dataset. Our pipeline, on the other hand, learns mask prediction in a weakly supervised way due to the stacked network paradigm  we adopt in MGBG (meeting criteria ii)). We train using wide/shallow depth-of-field images using existing datasets without mask ground-truth. The level of mask guidance we achieve is visually pleasant (see Figure \ref{fig:first_page}), it empowers users with greater flexibility (i.e. mask editing, providing custom masks) and even extends the use case beyond bokeh rendering to mask-based, learnable image refocusing. Note that our bokeh generation can still be guided with depth and/or saliency maps. Our mask prediction network also brings the controllable blur strength feature; by changing the background intensities of our guidance mask, we show that we can simulate different f-stops (meeting criteria iii), see Figure \ref{fig:fstops}). Our main contributions can be summarized as follows:

\begin{figure}[!t]
\includegraphics[width=\textwidth]{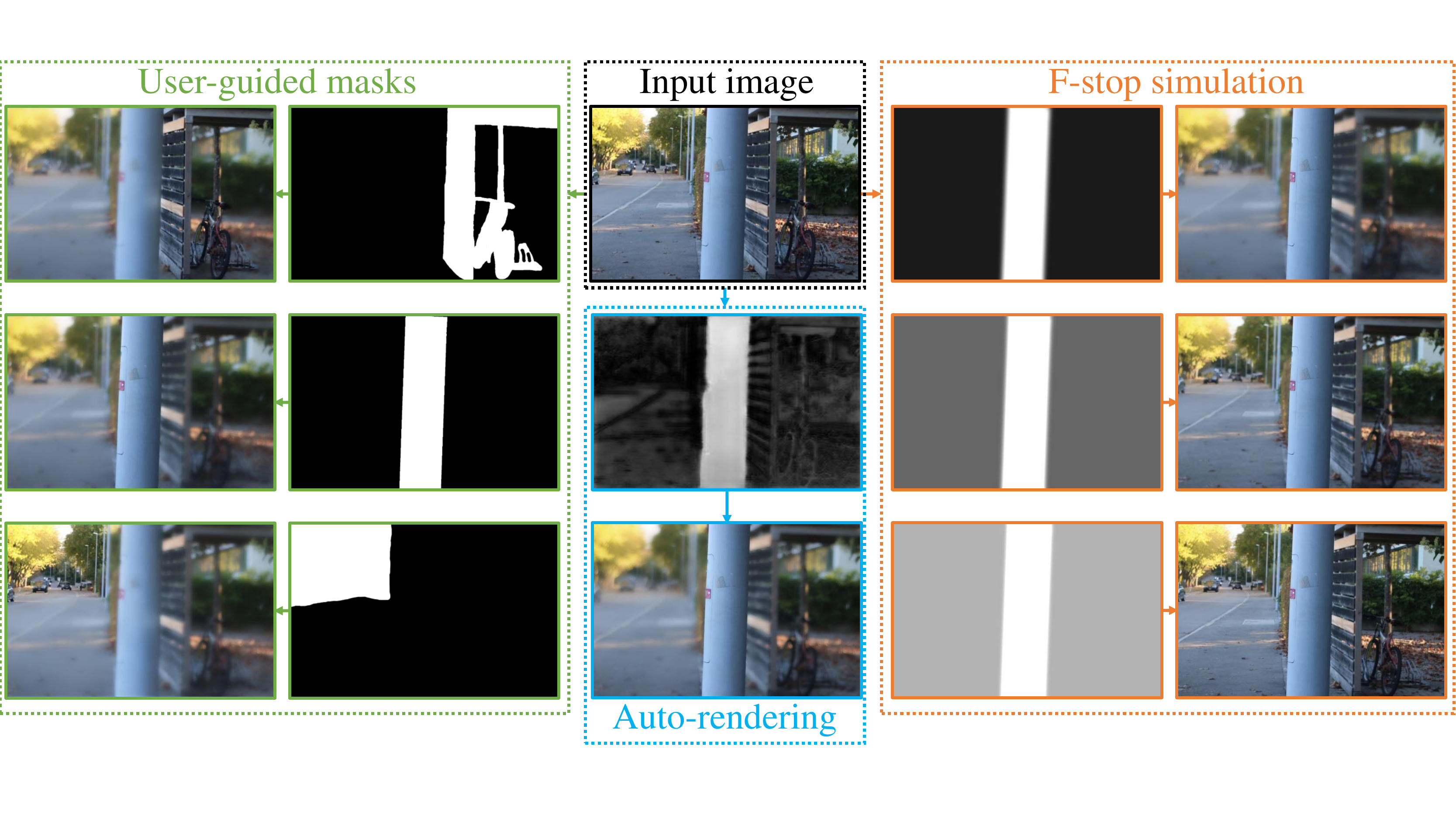}
\vspace{-14mm}
\caption{Our AMPN method renders bokeh images conditioned on masks, which can be provided by users and thus our results can focus on selected area of an image (left side). Furthermore, by changing the intensity of the guidance masks, we can simulate different f-stops (right side). Our weakly-supervised mask prediction network learns masks without direct supervision (middle images), which can be replaced by user-provided masks in inference.}
\label{fig:first_page}
\end{figure}

\begin{enumerate}
\vspace{-2mm}\item We propose AMPN for realistic bokeh rendering, which renders bokeh at low resolution via the MGBG block and performs refinement/upsampling with the LPR block. AMPN produces competitive or better results compared to existing methods on EBB!val294 set and performs faster than alternatives, making it ideal for mobile device deployment.
\item We show that the design of AMPN enables training of a mask prediction network in a weakly-supervised way, which lets users control their bokeh rendering with a mask; either obtained by the mask prediction network, edited by the user or even produced by the user.
\item We also show that by simply changing the intensity values of the guidance mask, we can control the strength of the blur in the bokeh rendered image.
\end{enumerate}

\section{Related work}

\noindent \textbf{Synthetic Bokeh Rendering.} Synthetic bokeh rendering can be largely divided into two categories \cite{peng2022bokehme} as classical and ML-based approaches. Classical rendering is shown to provide a degree of control and flexibility, but relies heavily on availability of accurate 3D information \cite{zhang2019synthetic,wadhwa2018synthetic,busam2019sterefo}. Here, we focus on the ML-based approaches due to their relevance to our method.

Following the portrait based methods \cite{shen2016automatic,shen2016deep}, more recent methods utilize prior knowledge, such as depth or saliency maps, along with the in-focus input image for high-fidelity bokeh rendering. In \cite{ignatov2020rendering}, the authors use an inverted pyramid-shaped architecture, which processes images at different scales, thus learning more diverse features. Their approach relies on depth estimation maps, produced by a pretrained Megadepth model \cite{li2018megadepth}. In \cite{purohit2019depth}, saliency maps, as well as depth maps are used for a spatially-aware blurring process. The maps are concatenated with the output of a space-to-depth module, before being fed to a densely connected encoder/decoder network that produces the bokeh image.

Parallel lines of work try to avoid depending on external information such as depth or saliency maps. \cite{choi2020efficient} proposes a fast generator architecture with a feature pyramid network and two discriminators, operating at global and patch levels. The authors of \cite{dutta2021stacked} propose a fast, stacked multi-scale hierarchical network that process images at different scales to obtain local and global features. In \cite{luo2020bokeh}, the authors propose a multi-module network, where each module focus on separate components such as defocus estimation, radiance, rendering and upsampling. \cite{qian2020bggan} uses a fast Glass-Net generator and Multi-Receptive-Field discriminator, and also re-implement instance normalization layers for further speed-ups on a mobile device. Authors of \cite{nagasubramaniam2022bokeh} propose a transformer-based architecture for bokeh rendering. They show that the increased receptive field, as well as pretraining the model on image restoration tasks, help improve the results. A recent work \cite{peng2022bokehme} propose a combination of classical and neural rendering, to achieve high-resolution photo-realistic bokeh images. Their framework is adjustable, where the blur size, focal plane and aperture shape can be chosen. Our approach combines the flexibility (i.e. user-editable mask guided bokeh rendering, varying blur strength) and mobile-friendliness of existing methods.

\noindent \textbf{Operating at High-resolutions.} A key aspect of mobile vision tasks, such as  dense regression/prediction (i.e. bokeh rendering, image editing, etc), is that they have a low tolerance to visual artefacts. Especially with the continuously improving mobile display technology, high-resolution images are becoming the norm, which makes the error tolerance even lower and requires accurate, high-resolution outputs. However, simply operating at high-resolution is not an option, especially due to the limited resources available on mobile devices. 

A naive approach is to perform the vision task at low resolution and then perform upsampling/super-resolution \cite{li2022ntire}. However, this approach requires a second model, which should work well on the same distribution and has to work sequentially with the vision model, which lowers the overall feasibility. Authors of \cite{liang2021high} propose the LPTN framework where upsampling and the vision task is performed jointly, where low-frequency components are translated with a lightweight network and high-frequency details are refined using both low and high-frequency components. Our LPR block is based on LPTN and its successors \cite{lei2022abpn}, with key differences explained later in relevant sections.

\begin{figure*}[!ht]
\includegraphics[width=\textwidth]{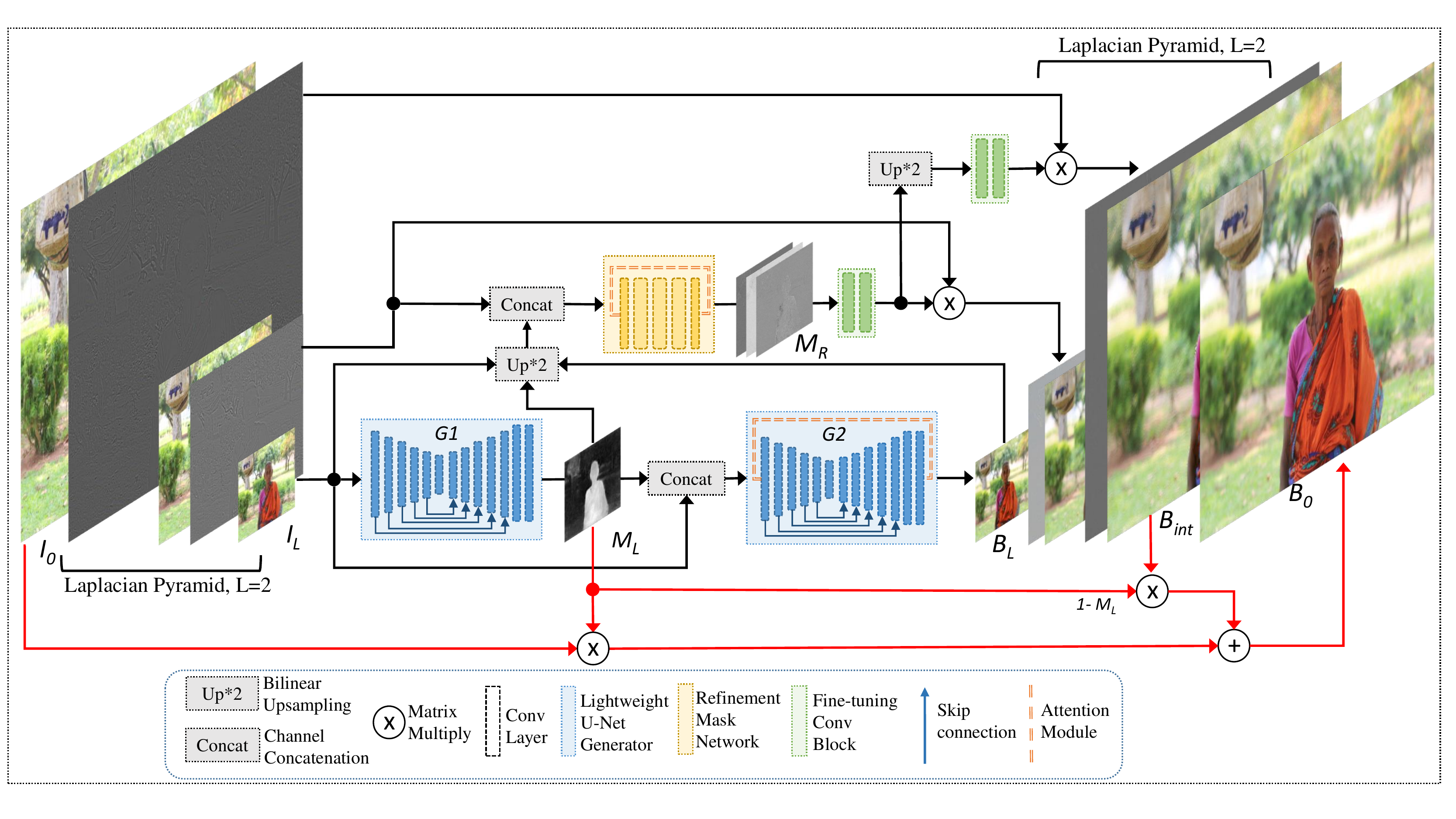}
\vspace{-10mm}
\caption{The diagram of our proposed AMPN pipeline. First, we decompose the input image $I_0$ with a Laplacian Pyramid and acquire the low-resolution image $I_L$. Through our MGBG  (light blue blocks) block, we first (in a weakly-supervised manner) discover the area of focus $M_L$ with $G1$, and then in $G2$ we use this area of focus to guide the bokeh rendering in low-resolution to produce $B_L$. Within our LRP block (refinement mask network and fine-tuning conv blocks), we progressively upsample and refine the low-resolution bokeh image and use mask-based blending (red arrows) to produce the final high-resolution bokeh image $B_0$. }
\label{fig:architecture}
\end{figure*}

\vspace{-5mm}

\section{Adaptive Mask-based Pyramid Network}
In this section, we motivate the need for a new bokeh rendering approach, introduce AMPN and its components in detail.

\subsection{Motivation}
Our aim to design a pipeline that meets the three criteria mentioned in earlier sections; i) fast processing with high-resolution outputs, ii) user-editable mask guided bokeh rendering and iii) ability to edit blur strength. We note that bokeh, by definition, is guided by depth information. Therefore, achieving criteria ii) provides a functionality beyond bokeh rendering, which is essentially image refocusing guided by a mask. In a sense, our aim is to learn from bokeh rendering datasets while being able to perform the more abstract task of mask-based image refocusing. Furthermore, achieving iii) makes our bokeh rendering hardware-independent; essentially we will not overfit to a specific bokeh style \cite{peng2022bokehme}. To summarize, our aim is beyond having an accurate, fast bokeh rendering model, but also to produce a highly interactive bokeh rendering experience.

\subsection{Overview}
Our AMPN pipeline is composed of two main blocks; the Mask-Guided Bokeh Generator (MGBG) block and a Laplacian Pyramid Refinement (LPR) block. The input image $I_0$ is decomposed, through a Laplacian pyramid, into its high-frequency components, denoted by $H=[h_0,h_1,...,h_{L-1}]$, and its low-frequency residual image $I_L$, where $L$ refers to the levels in the Laplacian pyramid \cite{liang2021high}. $I_L$ is then fed into the MGBG block, which generates the low-resolution mask $M_L$ and then the low-resolution bokeh image $B_L$. Next, $I_L$, $M_L$ and $B_L$ are upsampled with bilinear interpolation to match the size of the lowest size high-frequency component $h_{L-1}$, and then the four of them are concatenated and used as input to the LPR block. LPR generates a refinement mask $M_R$, which is combined progressively with the high frequency components $H$ of each level, producing the high-resolution image $B_{int}$. $B_{int}$ and $I_0$ are processed with $M_L$ to produce the final bokeh image $B_0$. The entire network is trained end-to-end, using high-resolution inputs and outputs. The overall diagram of AMPN is shown in Figure \ref{fig:architecture}. We now explain in more detail how the MGBG and LPRL blocks work and how they are trained.

\subsection{Mask-Guided Bokeh Generator Block}

The goal of the MGBG block is to generate the bokeh image in low-resolution. A natural approach here would be to use a single network that takes in the input image and simply produce the output image (i.e. optionally by using additional information such as depth and saliency maps). Since our aim is to also achieve criteria ii), we use a stacked-network formation using two networks, which we call $G1$ and $G2$. For both $G1$ and $G2$, we use accurate and performant architectures (criteria i)); we leverage the architecture of \cite{yucel2021real} which is a combination of MobileNetv2 \cite{sandler2018mobilenetv2} and FBNet \cite{wu2019fbnet} components.

$G1$ and $G2$ have their own separate tasks; $G1$ takes as input the input image $I_L$ and outputs a grayscale mask $M_L$, whereas $G2$ takes in $I_L$ and $M_L$ as input (as a 4-channel tensor) and generates the low-resolution bokeh rendered image $B_L$. We note that there is no ground-truth for $M_L$, therefore $G1$ is trained without any explicit supervision. $G1$, by leveraging the paired training samples, discovers the areas of refocus. The core advantages of having $G1$ are threefold: first, unlike other methods that do not use any guidance, $G1$ makes our network interpretable as it learns the area of refocus in the form of a mask  (see Figure \ref{fig:qualitative}). Second, since $G2$ is conditioned on the output of $G1$, we can control the refocus area (criteria ii), see Figure \ref{fig:first_page}). Finally, we can even remove $G1$ in inference and ask the user to provide any type of mask; user-generated, depth, saliency, etc (see Figure \ref{fig:masks}). This makes our method lightweight and applicable beyond bokeh rendering (i.e. image refocusing).

\begin{figure}[!t]
\includegraphics[width=\textwidth]{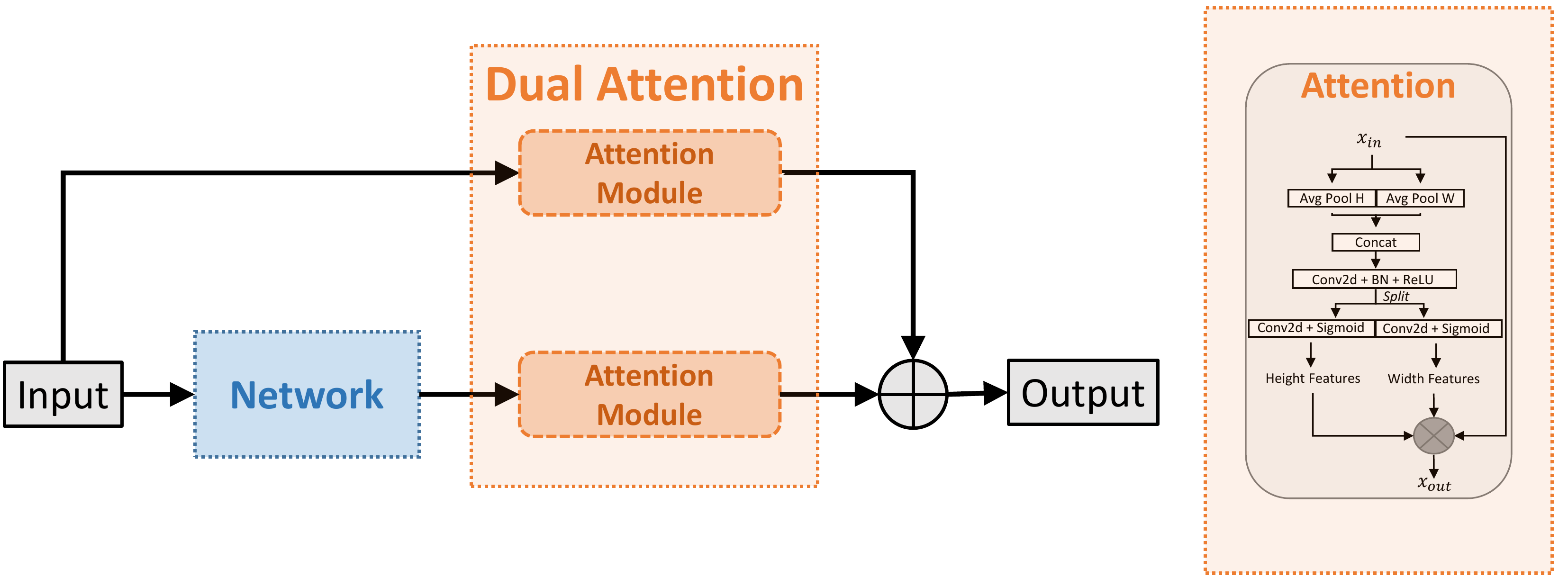}
\vspace{-5mm}
\caption{Our dual attention module. The \textit{network} block represents $G2$ network of the MBGB block, as well as the refinement network of the LPR block. The attention module is implemented via \cite{hou2021coordinate}, details of which are shown on the right. }
\label{fig:attention}
\end{figure}

The 4-channel input to $G2$ ($I_L$ and $M_L$) is passed through the $G2$ network that predicts an RGB image $I_{int}$. We also leverage a dual-attention mechanism that operates over the RGB component of the input (i.e. $I_L$) and $I_{int}$ (see Figure \ref{fig:attention}). Separate attention modules process $I_L$ and $I_{int}$, and their results are summed to produce the low-resolution bokeh image $B_L$. The inspiration for this dual-attention mechanism comes from modern super-resolution approaches \cite{ignatov2021real}, where long-distance residual connections are used. Instead of using residual connections directly from $I_L$, we leverage attention modules for visually pleasing results. We use \cite{hou2021coordinate} to implement the dual-attention modules. Note that each attention module is learned jointly with the entire pipeline, and they do not share weights. We also note that similar to $G1$, $G2$ is learned without direct supervision. We enforce our losses at the final, high-resolution bokeh image $B_0$ (i.e. the output of the LPR block).

\subsection{Laplacian Pyramid Refinement Block}
The goal of the LPR block is twofold; upsample the low-resolution bokeh rendered image $B_L$ back to the original resolution and refine/improve the results while doing so. Our LPR block is based on \cite{liang2021high}, with several key differences.

\noindent \textbf{Preliminaries.} The Laplacian pyramid \cite{burt1987laplacian} decomposes the image into low and high frequency components, from which the original image can be reconstructed. At each level of the pyramid, a fixed kernel is used to calculate the weighted average of the neighbouring pixels of the image, resulting in a downsampled version. The downsampled version is then upsampled again, and a high-frequency residual component $h$ is calculated by subtracting the upsampled image from the original one. This process is repeated for each pyramid level, and with the inverse operation, the original image is reconstructed. For each level, we refer to high-frequency residuals as $H=[h_0,h_1,...,h_{L-1}]$.

\noindent \textbf{LPR.} Having produced $M_L$ and $B_L$ via the MGBG block, we now aim to upsample $B_L$ to produce our final results $B_0$.  We first take $I_L$, $M_L$ and $B_L$ and upsample them to reach the spatial resolution of the lowest (pyramid level 2) high-frequency residual $h_{L-1}$, and then concatenate these four. This concatenated tensor is fed to the refinement network, which produces the refinement mask $M_R$. We note that we use the same dual-attention mechanism used in G2 (see Figure \ref{fig:attention}) in the refinement network as well; input attention processes the $B_L$ and the refinement network outputs an RGB image, which are then summed to produce $M_R$. $M_R$ is then processed with fine-tuning convolutional blocks (formed of two convolutional layers, and a LeakyReLu in between), and multiplied with $h_{L-1}$ to produce the output high-frequency residual of the first level. The same process is repeated for every level with the upsampled $M_R$ as the input, except we use the refinement network only on the first level. In LPR, we use a 2-level Laplacian pyramid, however we note that this can be extended to any number of levels depending on the capacity/output-resolution requirements. At the end, we end up with two output high-frequency residuals, which are added to progressively upsampled $B_L$ in each level, until we produce $B_{int}$ in the original resolution. As the final step, we take in the original high-resolution input $I_0$ and multiply it with $M_L$, and sum its results with $B_{int}$ multiplied with $1-M_L$ to produce the final bokeh rendered image $B_0$. We use masking so that our entire pipeline can focus on where matters (i.e. non-focus areas), since the input masking operation copies the focus area directly from the input image.  

\noindent \textbf{Differences with LPTN.} Our LPR block differs from LPTN in several aspects; i) In addition to $I_L$ and $B_L$, LPR also leverages $M_L$ in the upsampling, which guides the refinement/upsampling process, ii) we propose the dual-attention mechanism in the refinement network which improves it accuracy and iii) we use the mask $M_L$ to both leverage $I_0$ and better blend/integrate $I_0$ and $B_{int}$ spatially in generating the final output $B_0$.

\subsection{Losses} We use common regression losses, as well as perceptual-aware losses during our training. In total, we use three loss functions; the reconstruction loss $L_{1}$, the learned perceptual image patch similarity (LPIPS) \cite{zhang2018unreasonable} loss $L_{LPIPS}$ and the structural similarity loss $L_{SSIM}$ \cite{wang2004image}.  The final loss is a combination of the above losses, which is defined as: \[L_{total} = 10 \cdot L_{1} + 2 \cdot L_{LPIPS} + L_{SSIM}\]

The weights for each loss are based on \cite{qian2020bggan}. As noted before, the loss is applied only on the final bokeh image $B_0$, therefore $G1$ or $G2$ networks in the MGBG block do not have direct supervision. During our experiments, we experimented with applying the losses also over the output of $G2$ ($B_L$) and even over the outputs of each pyramid level (i.e. $B_{int}$), however, we did not see tangible improvements in either configuration.

\section{Experiments}

\subsection{Dataset}

We train our model on the \textit{Everything looks Better with Bokeh!} \textit{(EBB!)} dataset \cite{ignatov2020rendering}. It consists of 5K aligned image pairs of wide/shallow depth-of-field, 4600 of which are used as the training set and 200 images are used  for the validation and test sets, respectively. All the images in the dataset have a height of 1024 pixels, however, the width varies between images.

Since \textit{EBB!} is a challenge dataset \cite{ignatov2019aim,ignatov2020aim}, the ground-truths for validation and test sets are not publicly available. For a fair comparison, we use the \textit{val294} \cite{dutta2021depth} set for evaluation. The \textit{val294} set is based on the EBB!’s train set, where the first 4400 images are used for training, while the rest is used for evaluation.

\subsection{Experimental setup}

\textbf{Implementation Details.} We use PyTorch \cite{paszke2019pytorch} throughout our experiments. Following \cite{yucel2021real}, we initialize the encoders of $G1$ and $G2$ with weights obtained by training on ImageNet. The rest of the learnable parameters (i.e. LPR modules and the decoders of $G1$ and $G2$) are initialized with \cite{he2015delving}. All experiments are conducted on a PC with NVIDIA GeForce RTX 3090 GPU and AMD EPYC 7352 CPU. We train our model for 500 epochs, with a batch size of 8, utilizing the Adam optimizer and $2 \cdot 10^{-4}$ learning rate. The trainings are performed with input and output images with the size 1024x1536.

\noindent \textbf{Evaluation metrics.} We use Peak Signal-to-Noise Ratio (PSNR), Structural Similarity (SSIM) \cite{wang2004image} and Learned Perceptual Image Patch Similarity (LPIPS) \cite{zhang2018unreasonable} as our evaluation metrics. These metrics are the most widely used metrics used in the synthetic bokeh rendering literature, therefore we choose them to establish a fair comparison with existing methods. We note that many methods also leverage user surveys as another evaluation criteria; despite the value they bring, such surveys are generally not comparable nor reproducible. Furthermore, we believe the key contributions of our method, such as user-editable mask guidance and controllable blur strengths, can be well-represented with qualitative examples. Therefore, we leave user surveys as future work.

\subsection{Comparison with State-of-the-art}
We compare our method with several state-of-the-art methods, such as PyNet \cite{ignatov2020rendering}, DMSHN \cite{dutta2021stacked}, SKN \cite{li2019selective}, DBSI \cite{dutta2021depth} and BRViT \cite{nagasubramaniam2022bokeh}. We choose these methods based on the availability of their source codes and whether they report results on \textit{Val294} set on EBB!. We use pretrained models and source codes for evaluation (if available), or original results reported in relevant papers for other methods.

\noindent \textbf{Quantitative comparison.} The results are shown in Table \ref{tab:metrics}. Our method produces competitive results in each metric. In PSNR, our method trails behind others. Our method does quite well in SSIM and LPIPS metrics, it is the second best in SSIM and the best in LPIPS by a considerable margin. We note that BRViT (1st in SSIM) \cite{nagasubramaniam2022bokeh} uses quite a large model based on Vision Transformers, and also performs an initial pretraining stage. In general, our model performs competitively, and even outperforms others in perceptual metrics.

\noindent \textbf{Performance comparison.} We also compare our runtime performance against other state-of-the-art method. For a fair comparison, we only compare against methods that have publicly available source codes or pretrained models, so that we can compare them in our system in a fair fashion. The results are reported in Table \ref{tab:runtime}. In constrained scenarios, such as the desktop CPU, we outperform others with a significant margin. Furthermore, the last column shows that our method uses significantly fewer parameters. Finally, it is visible that our method has the fewest GFLOPs among other alternatives, proving its efficiency in terms of computation complexity. We note that our method is not likely to saturate high-end desktop GPUs, therefore it might not be as efficient as other larger methods that can utilize them better. Therefore, we highlight that our method is aimed at achieving good performance in resource-limited environments, such as mobile devices.

\begin{figure*}[!ht]
\includegraphics[width=\textwidth]{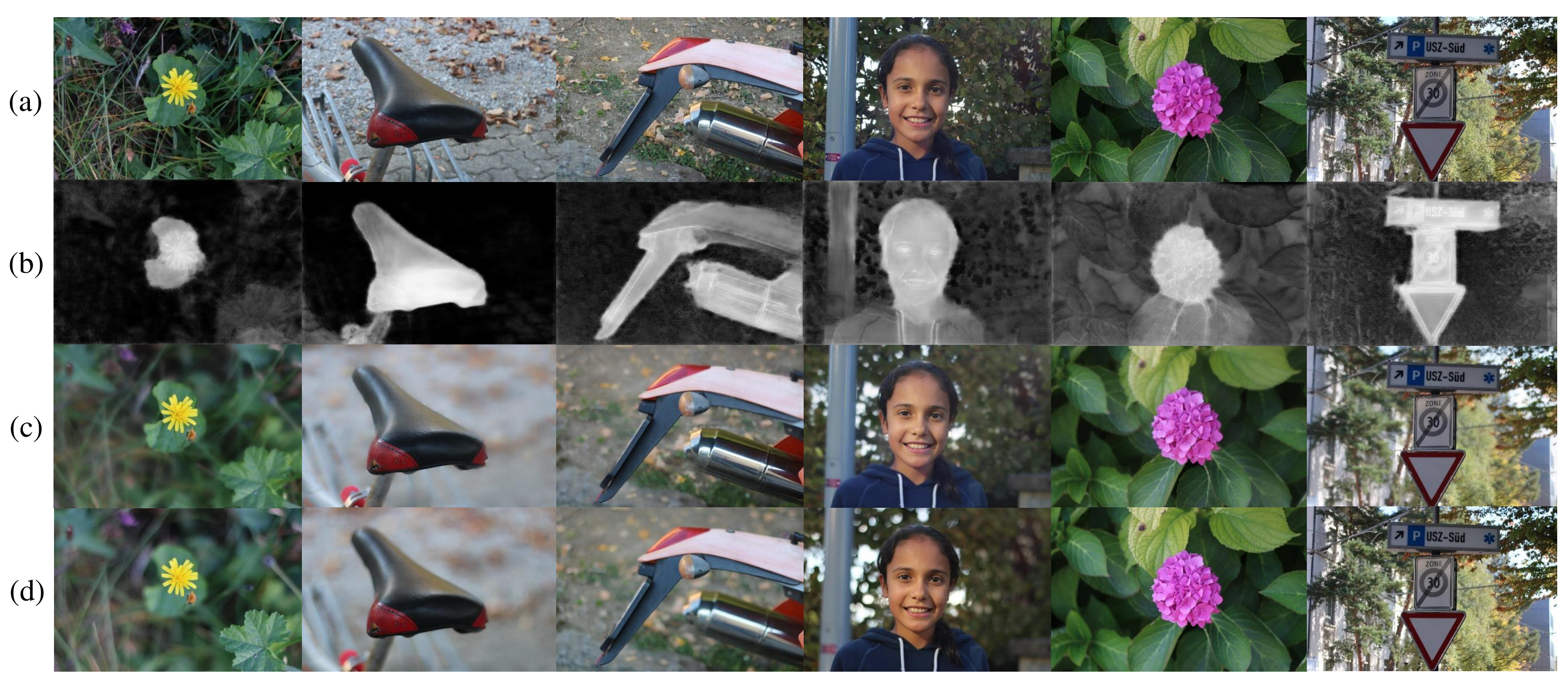}
\vspace{-5mm}
\caption{Qualitative results of our proposed method. (a) Input images, (b) $M_L$ masks predicted by $G1$ of the MGBG block, (c) rendered bokeh images and (d) ground-truth masks.}
\label{fig:qualitative_masks}
\end{figure*}

\begin{figure*}[!ht]
\includegraphics[width=\textwidth]{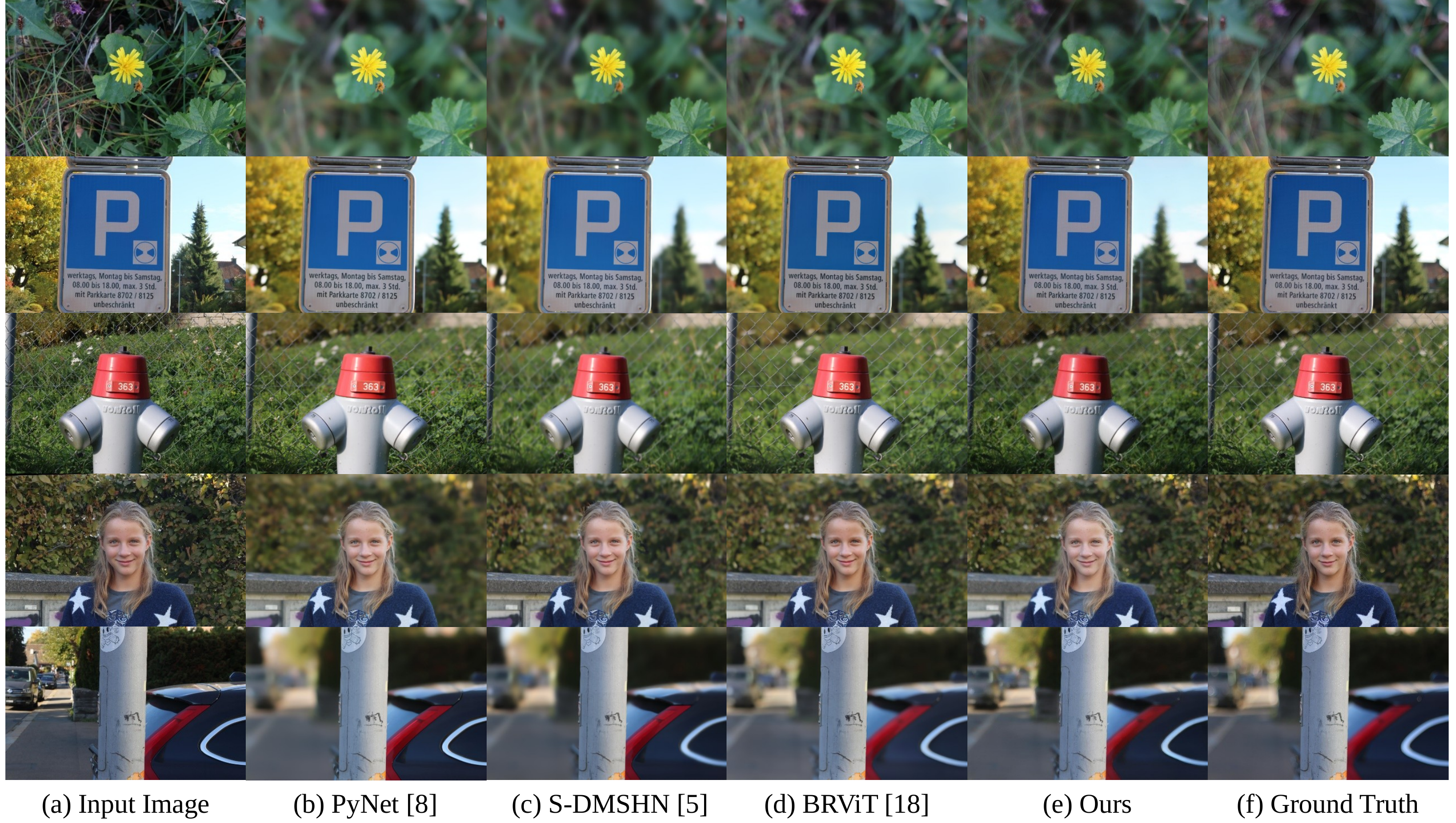}
\vspace{-5mm}
\caption{Qualitative comparison with other methods.}
\label{fig:qualitative}
\end{figure*}

\noindent \textbf{Qualitative comparison.} Example results produced by our method are shown in Figure \ref{fig:qualitative_masks}. $G1$ network in our MGBG module successfully learns, in a weakly supervised way, the areas of focus in the ground-truth data. Our learned masks are not strictly binary, which we later show to be useful for simulating different f-stops. Qualitative comparison against existing methods are shown in Figure \ref{fig:qualitative}. Our method performs competitively against the alternatives, producing visually pleasing images in multiple cases. We note that competing methods are slower and have significantly larger capacity compared to us, therefore our method might produce slightly worse results in some cases.

\begin{table}[!t]
\centering
\resizebox{0.5\textwidth}{!}{
	\begin{tabular}{c|c|c|c}
	
	Method & PSNR$\uparrow$ & SSIM$\uparrow$ & LPIPS$\downarrow$ \\ 
	\hline \hline
	SKN \cite{li2019selective} & 24.66 & 0.8521 & 0.3323 \\
	\hline
	DBSI \cite{dutta2021depth} & 23.45 & 0.8675 & 0.2463 \\ 
	\hline
	PyNet \cite{ignatov2020rendering} & \textbf{24.93} & 0.8788 & 0.2219 \\ 
	\hline 
	DMSHN \cite{dutta2021stacked} & 24.65 & 0.8765 & 0.2289 \\
	\hline
	Stacked DMSHN \cite{dutta2021stacked} & 24.72 & 0.8793 & 0.2271 \\ 
	\hline 
	BRViT \cite{nagasubramaniam2022bokeh} & 22.88 & 0.8516 & 0.2558 \\ 
	\hline \hline
	BRViT \cite{nagasubramaniam2022bokeh}  $\ddagger$ & \underline{24.76} & \textbf{0.8904} & \underline{0.1924} \\ 
	\hline \hline
	Ours & 24.50 & \underline{0.8847} & \textbf{0.1718} \\ 
	
	\end{tabular}
}
\vspace{2mm}
\caption{Comparison with other methods on \textit{Val294} set. SKN results are taken from \cite{nagasubramaniam2022bokeh}. BRViT \cite{nagasubramaniam2022bokeh} $\ddagger$ performs initial pretraining. The best and second best results are in bold and underlined, respectively, for each metric.} 
\label{tab:metrics}
\end{table}

\begin{table}[!t]
	\centering
\resizebox{0.60\textwidth}{!}{
	\begin{tabular}{c|c|c|c}
	Method &  CPU (sec) & \# of params & GFLOPs\\ 
	\hline \hline
	PyNet \cite{ignatov2020rendering} &  \underline{4.089} & 47.5M &  5300 \\ 
	\hline 
	Stacked DMSHN \cite{dutta2021stacked} & 12.657 & \underline{21.7M} & \underline{480.7} \\ 
	\hline 
	BRViT \cite{nagasubramaniam2022bokeh} & 30.288  & 123.1M &  650\\
	\hline \hline
	Ours & \textbf{1.591}  & \textbf{5.4M} &  \textbf{45.9} \\ 
	\end{tabular}
}
\caption{Performance comparison with other state-of-the-art methods, on a desktop CPU. Refer to the implementation details section for the evaluation setup.} 
\label{tab:runtime}
\end{table}

\subsection{Ablation studies}
\begin{figure}[!hb]
\includegraphics[width=\textwidth]{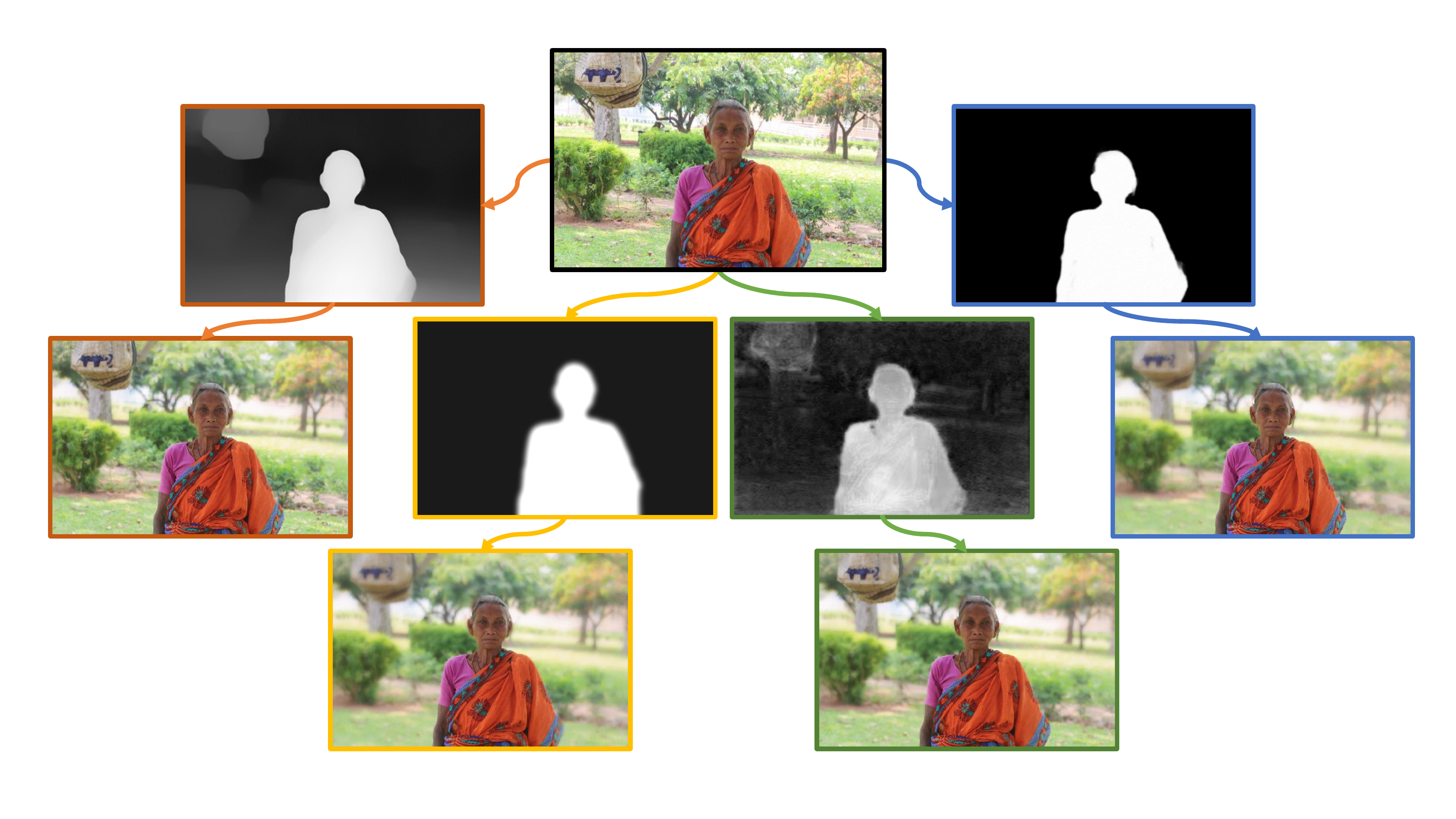}
\caption{Example outputs using different \textit{types} of masks; \textcolor{orange}{orange}, \textcolor{yellow}{yellow}, \textcolor{green}{green} and \textcolor{blue}{blue} masks are depth map, user-generated mask, $G1$-generated mask and saliency map, respectively.}
\vspace{-5mm}
\label{fig:masks}
\end{figure}

\textbf{Component Analyses.} We present the component analyses of our model architecture to show the contribution of the building blocks. The results are shown in Table \ref{tab:pipeline}. First, we test our pipeline without the refinement mask model in the LPR block. This variant (2nd row in Table \ref{tab:pipeline})  is clearly the fastest variant, on mobile GPU and desktop CPU alike, showing that the refinement mask network is a bottleneck in performance. However, we lose out quite a bit on LPIPS without the refinement mask model, which justifies its addition to our pipeline. Second, we remove $G2$ from the MGBG block; in this variant (3rd row in Table \ref{tab:pipeline}), we essentially force LPR to both learn bokeh rendering and upsample/refine the results. This variant does surprisingly well and performs slightly faster than the full pipeline. Third, we remove the dual attention mechanism from the refinement mask module in the LPR block. This variant (4th row in Table \ref{tab:pipeline}) is the close second in terms of evaluation metrics, and it is slightly faster than the full pipeline. Finally, we remove $G1$ from MGBG and simply use external masks to guide $G2$. This variant (5th row in Table \ref{tab:pipeline}) has similar runtimes with the 3rd variant. We note that this variant is a desirable configuration since external mask guidance is a more interactive use case, and also we can not report metrics with this variant since we use external masks. Finally, our full pipeline (1st row in Table \ref{tab:pipeline}) produces the best SSIM and LPIPS metrics, showing the effectiveness of our design choices.  We note that the other three variants shown in Table \ref{tab:pipeline}  (2nd, 3rd and 4th rows) are still competitive to other methods.

\begin{table}[!t]
\resizebox{\textwidth}{!}{
	\begin{tabular}{l|c|c|c|c|c|c|c}
	Components & PSNR$\uparrow$ & SSIM$\uparrow$ & LPIPS$\downarrow$ & D CPU (s) & D GPU (s) &  M GPU (s) & \# of params\\ 
	\hline \hline
	G1 + G2 + LPR  & \textbf{24.50} & \textbf{0.8847} & \underline{0.1718} & 1.591 & 0.059 & 0.330 & 5.4M\\
	\hline 
	G1 + G2 + LPR  (w/o ref.)  & 24.24 & 0.8814 & 0.1931 & \textbf{0.731} & 0.040 & \textbf{0.118} & 5.2M \\ 
	\hline
	G1, no G2 + LPR & 23.91 & 0.8692 & 0.1903 & 1.483 & 0.042 & 0.297 & 2.8M \\ 
	\hline
	G1 + G2 + LPR (w/o att.) & \underline{24.25} &  \underline{0.8825} & \textbf{0.1717} & 1.581 & 0.052 & 0.319  & 5.2M\\ 
	\hline
	no G1, G2 + LPR & \textbf{--} & \textbf{--} & \textbf{--} & 1.4732 & \textbf{0.039} & 0.312 & 2.8M \\ 
	\end{tabular}
}
\caption{Ablation study on our model pipeline on \textit{Val294} set. \textit{w/0 ref.} indicates LPR without the refinement mask model. \textit{w/0 att.} indicates LPR without dual attention in the refinement mask model. \textbf{D} and \textbf{M} stand for desktop and mobile runtimes, respectively. The desktop hardware is detailed in the implementation details section. Mobile runtimes are averaged on 50 runs, and acquired using Samsung Galaxy S22 Ultra with Exynos 2200 processor.} 
\label{tab:pipeline}
\vspace{-10mm}
\end{table}

\noindent \textbf{Using different masks.} As explained in earlier sections, our pipeline is flexible as it can be guided by any mask. This mask can be generated by $G1$ of the MGBG block, but also can be generated by a user. Furthermore, one can guide the bokeh generation with depth maps or saliency maps. We provide several scenarios where we generate the bokeh effect with different types of mask guidance, see Figure \ref{fig:masks} for examples. Our method manages to refocus images successfully with various types of mask guidance.

\begin{figure}[!ht]
\includegraphics[width=\textwidth]{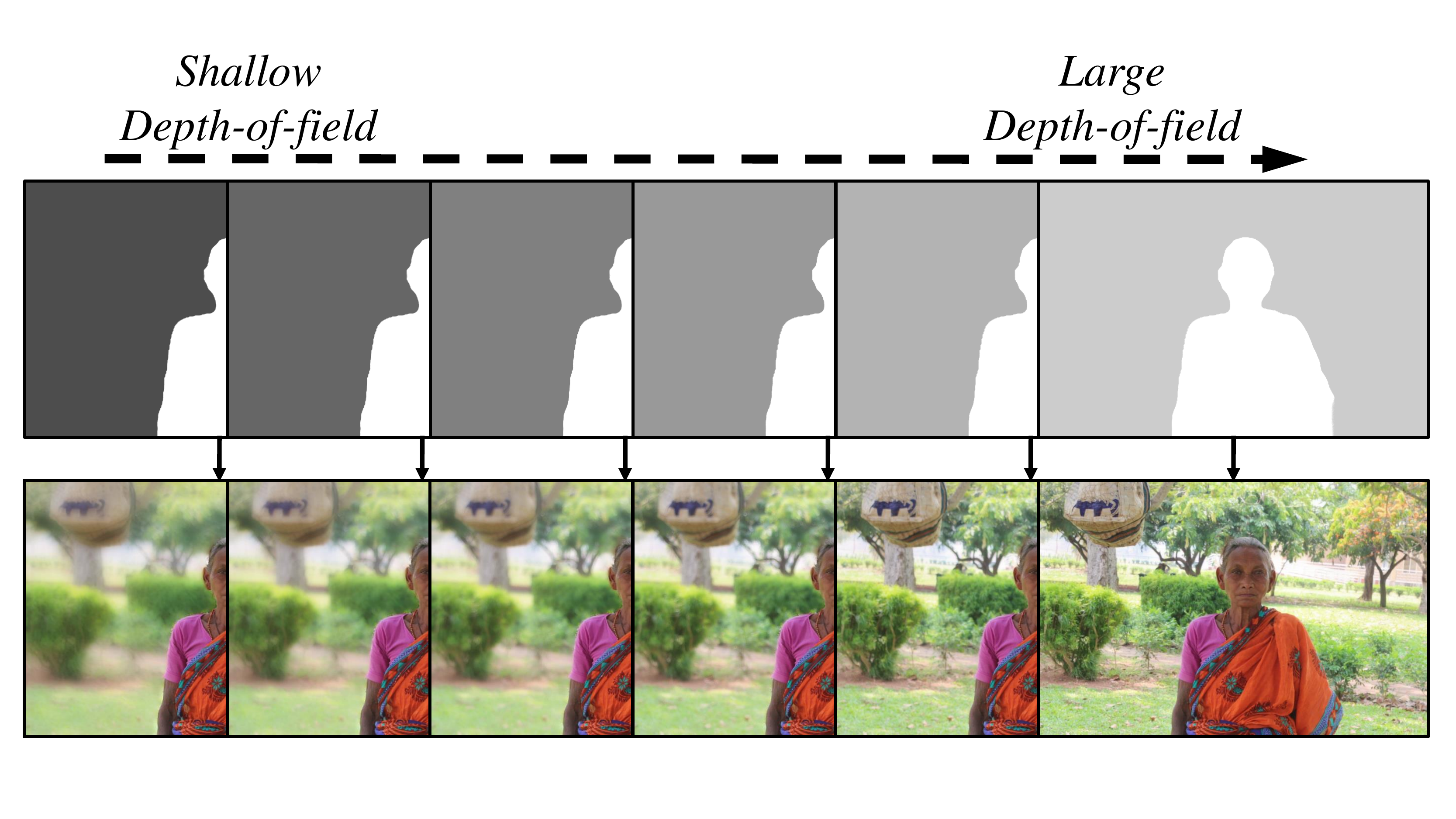}
\vspace{-14mm}
\caption{Different f-stop simulation. Each column shows the mask and the related bokeh rendering result. By changing the mask intensity values of the out-of-focus areas, we manage to simulate different f-stops. }
\label{fig:fstops}
\end{figure}

\noindent \textbf{Simulating different f-stops.} One of the important features of our method is that it can simulate different f-stops/blur strengths. We discover that by modifying the values of the input mask to our model ($G2$), we can achieve a certain level of control over various depth-of-focus effects. Specifically, we do not alter the mask intensities on the in-focus area, but rather alter the intensity values of the background (out-of-focus) areas. Coupled with the fact that we can use any mask to guide our bokeh/refocusing, this also provides another level of interactive experience, which further empowers the user. We provide several examples in Figure \ref{fig:fstops}.

\vspace{-3mm}

\vspace{-5mm}
\section{Conclusion}

In this paper, we focus on the task of ML-based synthetic bokeh rendering. We focus on three areas of improvement; i) accurate on-device performance with high-resolution images, ii) ability to guide bokeh-generation with user-editable masks and iii) ability to produce blurs with varying strength profiles. We propose the AMPN pipeline, which is formed of the MGBG and the LPR blocks. The MGBG block performs bokeh rendering at low resolutions for low memory footprint and fast processing, whereas LPR progressively upsamples and refines the low-resolution bokeh image generated by MGBG. Owing to its two-network design, the first network of MGBG discovers in-focus areas of images without supervision, in the form a mask. Since the second network of MGBG conditions the bokeh generation on this (weakly-supervised) discovered mask, we can edit/change the mask freely to guide bokeh generation. This level of control lets us achieve the more abstract functionality of mask-guided image refocusing. Furthermore, by changing the intensities of the mask, we show that we can simulate various blur strengths. Finally, we show that our method produces competitive or better results compared to various alternatives on the EBB! dataset, while running faster.

\bibliographystyle{splncs04}
\bibliography{egbib}
\end{document}